\documentclass[letterpaper,10pt,conference]{ieeeconf}

\IEEEoverridecommandlockouts
\usepackage{amsmath,amssymb}
\usepackage{graphicx}
\usepackage{xcolor}
\usepackage{listings}
\usepackage{microtype}
\usepackage{tikz}
\usepackage{url}
\usepackage[hidelinks]{hyperref}
\usetikzlibrary{arrows.meta,positioning,fit,backgrounds}

\definecolor{codeKeyword}{RGB}{45,72,128}
\definecolor{codeType}{RGB}{145,82,35}
\definecolor{codeComment}{RGB}{72,112,82}
\definecolor{codeString}{RGB}{120,75,105}
\definecolor{codeText}{RGB}{35,39,47}
\definecolor{codeBg}{RGB}{247,248,251}
\definecolor{codeRule}{RGB}{184,189,199}

\lstdefinelanguage{Rust}{
  sensitive=true,
  morekeywords={
    as,async,await,break,const,continue,crate,dyn,else,enum,extern,false,
    fn,for,if,impl,in,let,loop,match,mod,move,mut,pub,ref,return,self,
    Self,static,struct,super,trait,true,type,unsafe,use,where,while,assert
  },
  morecomment=[l]{//},
  morecomment=[s]{/*}{*/},
  morestring=[b]"
}

\title{\LARGE \bf
Cybflight: An Embedded Rust Autopilot for Aerial Robotics Research
}

\author{Yifan Lin$^{*}$, Chao Qin, H~S~Helson Go,
and Hugh H.-T. Liu
\thanks{The authors are with the Institute for Aerospace Studies,
University of Toronto, Toronto, Canada. Emails:
{\tt \{i.lin,\allowbreak{}chao.qin,\allowbreak{}hei.go\}\allowbreak{}@mail.utoronto.ca}
and {\tt hugh.liu\allowbreak{}@utoronto.ca}.}
\thanks{$^{*}$Corresponding author.}
}

\begin{document}

\maketitle
\thispagestyle{empty}
\pagestyle{empty}

\begin{abstract}
Bringing an aerial robotics method from simulation to flight should not require
rebuilding a mature autopilot or adding a companion computer. Cybflight is an
open-source embedded Rust research autopilot whose typed, replaceable interfaces
connect hardware access, perception, state estimation, trajectory planning,
and control. This modular development and compile-time optimization workflow is
demonstrated with replaceable Rust implementations of model predictive
contour-tracking control (MPCTC) and incremental nonlinear dynamic inversion
(INDI) running on one STM32H743 without a companion computer. Using this configuration, the vehicle reaches
12.38\,m/s during indoor flight, while an outdoor flight using global navigation satellite system
(GNSS) position updates reaches 31.4\,m/s. These flights show that running
demanding estimation and nonlinear control entirely on a flight-controller microcontroller need not come at the expense of a modular autopilot structure.
\par{Code:} \href{https://github.com/cybird-robotics/cybflight}{\texttt{github.com/cybird-robotics/cybflight}}
\par{Video:} \href{https://youtu.be/awRv8zCM5KA}{\texttt{youtu.be/awRv8zCM5KA}}
\end{abstract}

\section{Introduction}

Research on agile flight often starts from a mature autopilot that already
provides reliable operation and broad vehicle support. The challenge comes
when a core estimator or controller needs to be replaced: device
abstractions, configuration layers, and coupled subsystems can turn a local
change into substantial firmware-specific work. One can then spend as much
effort adapting the flight stack as evaluating the method.

A common workaround is to keep low-level control on the embedded flight
controller and move experimental autonomy to a Linux companion computer. This
provides more compute, but also adds mass, power, wiring, software integration,
and cost. These penalties matter on small high-speed aircraft, where payload
margin is scarce and each crash risks another expensive component.

To address these limitations, we built Cybflight around a simple goal: changing
the underlying hardware or replacing an estimator, controller, or learned policy
should not require reorganizing neighboring modules. Typed interfaces separate
hardware access, perception, state estimation, trajectory planning, control,
and actuation. A vehicle description selects concrete implementations at build
time, allowing the compiler to specialize a \texttt{no\_std} program for the
target microcontroller.

We demonstrate this workflow by implementing MPCTC and INDI as replaceable Rust
modules. Cybflight compiles them with the selected estimator and hardware
support into one program for a single flight-controller microcontroller.
High-speed indoor and outdoor flights then evaluate the resulting autopilot.
Our contributions are:
\begin{itemize}
  \item we introduce Cybflight, an open-source \texttt{no\_std} Rust autopilot
  that separates hardware-specific code from replaceable research algorithms;
  \item we demonstrate its modular development and compile-time optimization
  workflow through Rust implementations of MPCTC and INDI integrated into a
  single-microcontroller autopilot;
  \item we validate fully onboard autonomous flight indoors and outdoors at
  12.38\,m/s and 31.4\,m/s, respectively.
\end{itemize}

\begin{figure}
\centering
\includegraphics[width=\columnwidth,trim=300 0 300 0,clip]
  {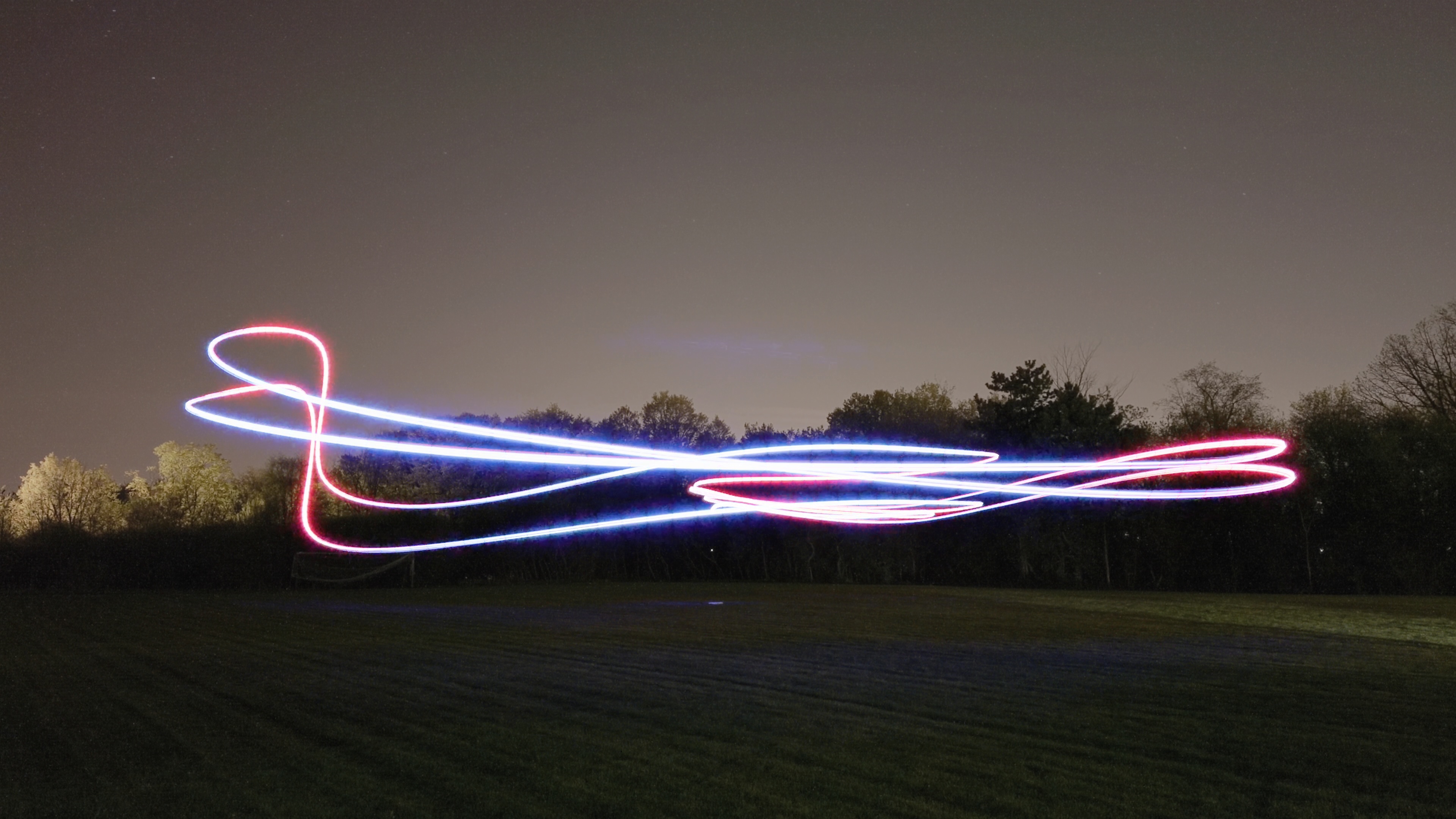}
\caption{Outdoor Split-S maneuver. The corresponding flight reached
31.4\,m/s with global navigation satellite system (GNSS) position updates and
fully onboard MPCTC--INDI.}
\label{fig:teaser}
\end{figure}

\section{Related Work}

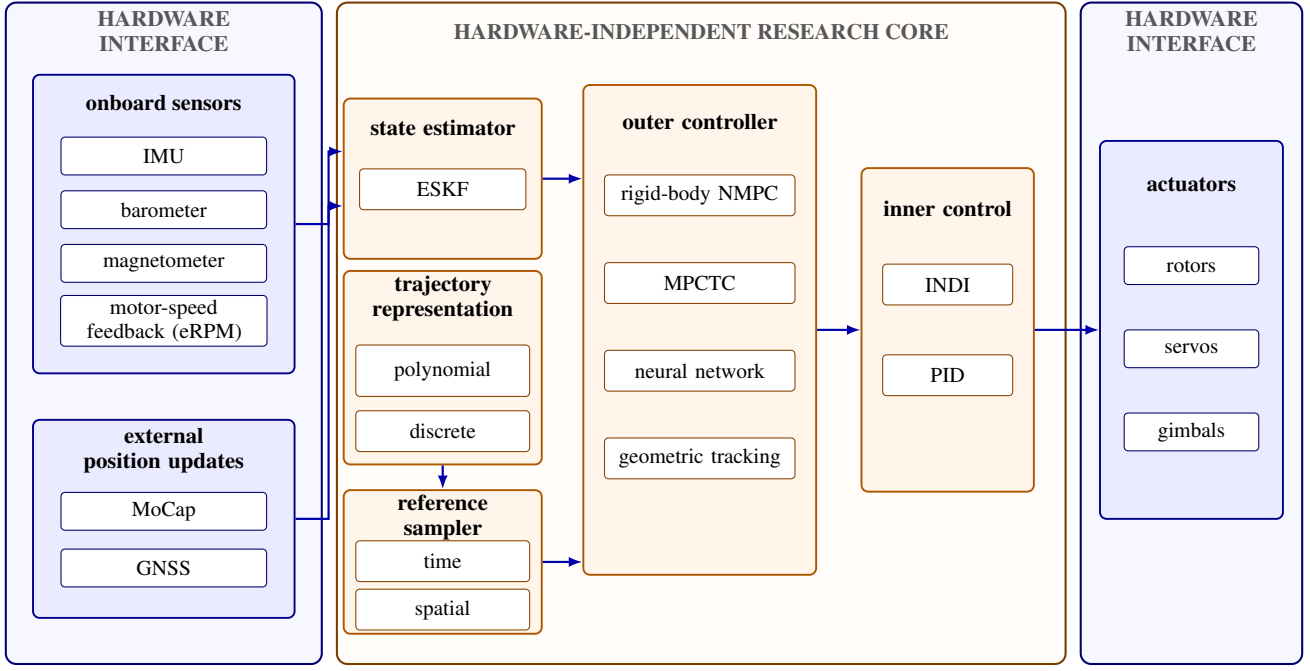
\begin{figure*}
\centering
\begin{tikzpicture}[
  x=1cm,y=1cm,
  font=\footnotesize,
  platform module/.style={draw=blue!55!black,rounded corners=2.5pt,
                          fill=blue!8,line width=0.72pt},
  core module/.style={draw=orange!68!black,rounded corners=2.5pt,
                      fill=orange!8,line width=0.72pt},
  platform item/.style={draw=blue!42!black,rounded corners=1.5pt,
                        fill=white,minimum height=0.50cm,align=center,
                        inner xsep=3pt,inner ysep=1.6pt,
                        font=\fontsize{8.0}{9.0}\selectfont},
  core item/.style={draw=orange!52!black,rounded corners=1.5pt,
                    fill=white,minimum height=0.54cm,align=center,
                    inner xsep=3pt,inner ysep=1.6pt,
                    font=\fontsize{8.0}{9.0}\selectfont},
  module title/.style={font=\fontsize{8.4}{9.4}\selectfont\bfseries,
                       inner sep=0pt,align=center},
  region title/.style={font=\fontsize{8.3}{9.3}\selectfont\bfseries,
                       inner sep=0pt,text=black!66,align=center},
  channel/.style={-{Latex[length=1.8mm]},draw=blue!67!black,
                  line width=0.82pt}
]

  \fill[blue!3,rounded corners=4pt] (0,0) rectangle (4.18,8.75);
  \draw[blue!44!black,rounded corners=4pt,line width=0.78pt]
        (0,0) rectangle (4.18,8.75);
  \fill[orange!2.5,rounded corners=4pt] (4.38,0) rectangle (14.02,8.75);
  \draw[orange!48!black,rounded corners=4pt,line width=0.78pt]
        (4.38,0) rectangle (14.02,8.75);
  \fill[blue!3,rounded corners=4pt] (14.22,0) rectangle (17.15,8.75);
  \draw[blue!44!black,rounded corners=4pt,line width=0.78pt]
        (14.22,0) rectangle (17.15,8.75);

  \node[region title,text width=2.55cm] at (2.09,8.37)
        {HARDWARE INTERFACE};
  \node[region title] at (9.20,8.37) {HARDWARE-INDEPENDENT RESEARCH CORE};
  \node[region title,text width=2.45cm] at (15.685,8.37)
        {HARDWARE INTERFACE};

  \node[platform module,minimum width=3.46cm,minimum height=3.95cm]
        (onboard) at (2.09,5.82) {};
  \node[module title] at (2.09,7.42) {onboard sensors};
  \node[platform item,minimum width=2.72cm] at (2.09,6.72) {IMU};
  \node[platform item,minimum width=2.72cm] at (2.09,6.01) {barometer};
  \node[platform item,minimum width=2.72cm] at (2.09,5.30) {magnetometer};
  \node[platform item,minimum width=2.72cm,text width=2.52cm]
        at (2.09,4.54) {motor-speed feedback (eRPM)};

  \node[platform module,minimum width=3.46cm,minimum height=2.62cm]
        (external) at (2.09,1.92) {};
  \node[module title,text width=3.0cm] at (2.09,2.83)
        {external position updates};
  \node[platform item,minimum width=2.72cm] at (2.09,2.02) {MoCap};
  \node[platform item,minimum width=2.72cm] at (2.09,1.27) {GNSS};

  \node[core module,minimum width=2.62cm,minimum height=2.12cm]
        (estimator) at (5.78,6.42) {};
  \node[module title,text width=2.05cm] at (5.78,7.10)
        {state estimator};
  \node[core item,minimum width=2.20cm] at (5.78,6.28) {ESKF};

  \node[core module,minimum width=2.62cm,minimum height=2.56cm]
        (trajectory) at (5.78,3.92) {};
  \node[module title,text width=2.20cm] at (5.78,4.84)
        {trajectory\\representation};
  \node[core item,minimum width=2.30cm,
        minimum height=0.68cm] at (5.78,3.88)
        {polynomial};
  \node[core item,minimum width=2.30cm] at (5.78,3.08)
        {discrete};

  \node[core module,minimum width=2.62cm,minimum height=1.90cm]
        (sampler) at (5.78,1.35) {};
  \node[module title,text width=2.20cm] at (5.78,1.95)
        {reference sampler};
  \node[core item,minimum width=2.30cm] at (5.78,1.36)
        {time};
  \node[core item,minimum width=2.30cm] at (5.78,0.72)
        {spatial};

  \node[core module,minimum width=3.08cm,minimum height=6.48cm]
        (outer) at (9.18,4.42) {};
  \node[module title,text width=2.35cm] at (9.18,7.17)
        {outer controller};
  \node[core item,minimum width=2.52cm] at (9.18,6.20)
        {rigid-body NMPC};
  \node[core item,minimum width=2.52cm] at (9.18,5.04) {MPCTC};
  \node[core item,minimum width=2.52cm] at (9.18,3.88) {neural network};
  \node[core item,minimum width=2.52cm] at (9.18,2.72)
        {geometric tracking};

  \node[core module,minimum width=2.28cm,minimum height=4.28cm]
        (inner) at (12.46,4.42) {};
  \node[module title,text width=1.95cm] at (12.46,6.03)
        {inner control};
  \node[core item,minimum width=1.72cm] at (12.46,5.02) {INDI};
  \node[core item,minimum width=1.72cm] at (12.46,3.82) {PID};

  \node[platform module,minimum width=2.42cm,minimum height=5.00cm]
        (actuator) at (15.685,4.42) {};
  \node[module title] at (15.685,6.35) {actuators};
  \node[platform item,minimum width=1.78cm] at (15.685,5.27) {rotors};
  \node[platform item,minimum width=1.78cm] at (15.685,4.17) {servos};
  \node[platform item,minimum width=1.78cm] at (15.685,3.07) {gimbals};

  \draw[channel] (onboard.east) -- (4.27,5.82)
       |- ([yshift=0.36cm]estimator.west);
  \draw[channel] (external.east) -- (4.27,1.92)
       |- ([yshift=-0.36cm]estimator.west);
  \draw[channel] (estimator.east) -- (outer.west |- estimator.east);
  \draw[channel] (trajectory.south) -- (sampler.north);
  \draw[channel] (sampler.east) -- (outer.west |- sampler.east);
  \draw[channel] (outer.east) -- (inner.west);
  \draw[channel] (inner.east) -- (actuator.west);

\end{tikzpicture}
\caption{Cybflight's flight graph. Hardware interfaces (blue) map sensor
observations and actuator commands to a hardware-independent research core
(orange). Stacked entries denote sensor and actuator classes or algorithm
families admitted by the interfaces; a vehicle build selects a compatible
estimation, trajectory representation, sampling policy, and control path.}
\label{fig:architecture}
\end{figure*}

To place Cybflight in context, we ask how existing platforms admit a new
algorithm and whether it runs on the flight-controller microcontroller, a
companion computer, or offboard hardware. That choice directly shapes the
research workflow.

PX4 provides embedded modularity through NuttX tasks and uORB
publish--subscribe middleware~\cite{meier2015px4}, however in practice, adding a new estimator or controller often requires firmware-specific work
\cite{dangelo2024px4}. Researchers often use Betaflight
as a high-rate inner-loop controller behind a software bridge; in
\cite{bosello2024race}, Robot Operating System 2 (ROS~2) autonomy runs on an
NVIDIA Jetson Orin NX and exchanges inertial measurement unit (IMU) data and
radio-control channels through Betaflight's MultiWii Serial Protocol override
mode. MonoRace uses a related split: vision and the main estimator remain on a
Jetson Orin NX, while a direct-to-motor neural policy runs on an
embedded microcontroller~\cite{bahnam2026monorace,blaha2024indiflight}.
Agilicious co-designs open hardware and software around interchangeable
estimators, controllers, and flight-controller bridges~\cite{foehn2022agilicious}.

Model predictive control (MPC) delivers high-performance flight by considering vehicle dynamics and constraints, but its online optimization is
computationally demanding. As a result, model predictive contouring control
(MPCC) for aggressive flight typically runs on a companion
computer~\cite{romero2022mpcc}: MPCC++ uses a 100-Hz optimizer on an Intel
processor and sends thrust and body-rate commands to
Betaflight~\cite{krinner2024mpccpp}. Recent work has brought MPC
onto microcontrollers by narrowing the online problem. TinyMPC uses linear
dynamics~\cite{nguyen2024tinympc}, while embedded nonlinear model predictive
control (NMPC) on nano quadrotors uses short horizons and real-time
iterations~\cite{kazim2023embedded}.

Rust is also gaining ground in aerial robotics through incremental C/C++
migration, shared embedded/host estimation, and educational flight
stacks~\cite{engels2025forces,schulz2025rio,honig2026rusty}. Cybflight brings
these strands together in one Rust-native embedded program: replaceable
interfaces span the perception-to-actuation path, while state estimation, a
20-step nonlinear MPCTC, and INDI run on the same microcontroller.

\section{Cybflight Architecture}

Cybflight combines two design choices: modular interfaces separate hardware
support from replaceable research algorithms, and an embedded numerical core
keeps model-based code readable and predictable on the microcontroller.

\subsection{Separating hardware from research algorithms}

Fig.~\ref{fig:architecture} shows how typed hardware interfaces isolate
hardware-facing perception and actuation from a board-independent research
core. The key idea is to exchange information rather than device handles. Typed
observations, vehicle states, references, and actuator commands form the module
boundaries. Motion capture (MoCap) and GNSS adapters publish the same
vehicle-odometry type, allowing us to change the position source without
changing estimation or control. We also separate trajectory representation
from reference sampling. At build time, a vehicle configuration selects
concrete implementations, allowing the compiler to specialize the resulting
code; fixed-capacity channels and workspaces make memory use known before
flight. These boundaries let us replace a board, estimator, controller, or
learned policy without restructuring neighboring modules.

\subsection{Embedded numerical core}

\begin{lstlisting}[caption={Fixed-size attitude Hessian shared by MPCTC and NMPC.},label={lst:quat-hessian}]
const { assert!(NX >= 7); }
let j: SMatrix<f32, 3, 4> = de * dqa_dq;
let w = Matrix3::from_diagonal(
    &Vector3::from(*w_att));
let h: SMatrix<f32, 4, 4> =
    j.transpose() * w * j * (2.0 * dt);
hess_xx.fixed_view_mut::<4, 4>(3, 3)
    .copy_from(&h);
\end{lstlisting}

For researchers to develop and modify algorithms rapidly, the implementation
must keep the underlying mathematics readable. We use fixed-size
\texttt{nalgebra}~\cite{crozet2026nalgebra} types in
\texttt{no\_std} mode, keeping vectors, matrices, rotations, and unit
quaternions explicit without dynamic allocation. The compiler can then catch
dimension mismatches in code that stays close to the equations.
Listing~\ref{lst:quat-hessian} shows the Gauss--Newton attitude Hessian shared
by the MPCTC and NMPC implementation; the operand types
encode its matrix dimensions. We use the same pattern for model Jacobians, cost derivatives, and horizon
workspaces.

\section{High-Speed Flight Validation}

We test Cybflight indoors and outdoors. In both settings, we run
the complete perception-to-actuation path onboard and change only the position
source. High-speed flight provides a demanding system-level test beyond a
comparison between control methods.

\subsection{Experimental configuration}

\begin{figure}
\centering
\includegraphics[width=\columnwidth,trim=0 84bp 0 0,clip]
  {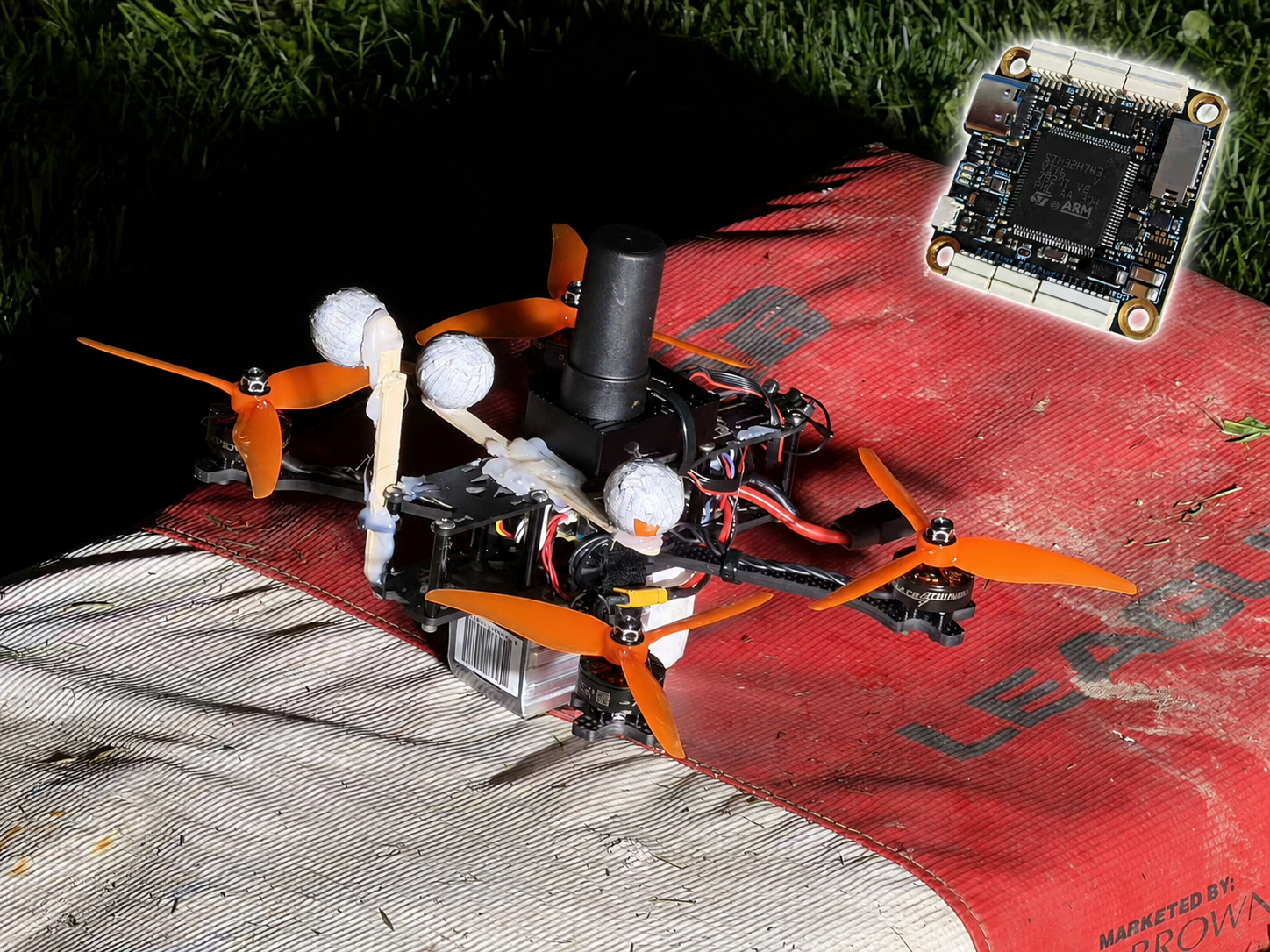}
\caption{Experimental quadrotor and STM32H743 flight controller (upper right),
which executes all estimation and control.}
\label{fig:vehicle}
\end{figure}

Unless otherwise stated, we select an error-state Kalman filter
(ESKF), MPCTC, and an INDI inner loop. MPCTC commands collective thrust and body
rates; INDI tracks them using angular acceleration and a
control-effectiveness model before allocating rotor
commands~\cite{smeur2016indi}. We generate the reference using our
previous planning methods~\cite{qin2024togt,qin2026ijrr}.

The vehicle in Fig.~\ref{fig:vehicle} has a mass of 0.60\,kg, a 0.227-m
diagonal arm length, and a nominal thrust-to-weight ratio of 8.2. We run the
complete feedback path on a 480-MHz STM32H743 without a companion computer.
MoCap and real-time kinematic (RTK) GNSS update the ESKF indoors and outdoors,
respectively. We define
tracking error as the three-dimensional distance from estimated position to the
nearest point on the planned path.

\subsection{Indoor validation}

\begin{figure}
\centering
\includegraphics[width=\columnwidth]{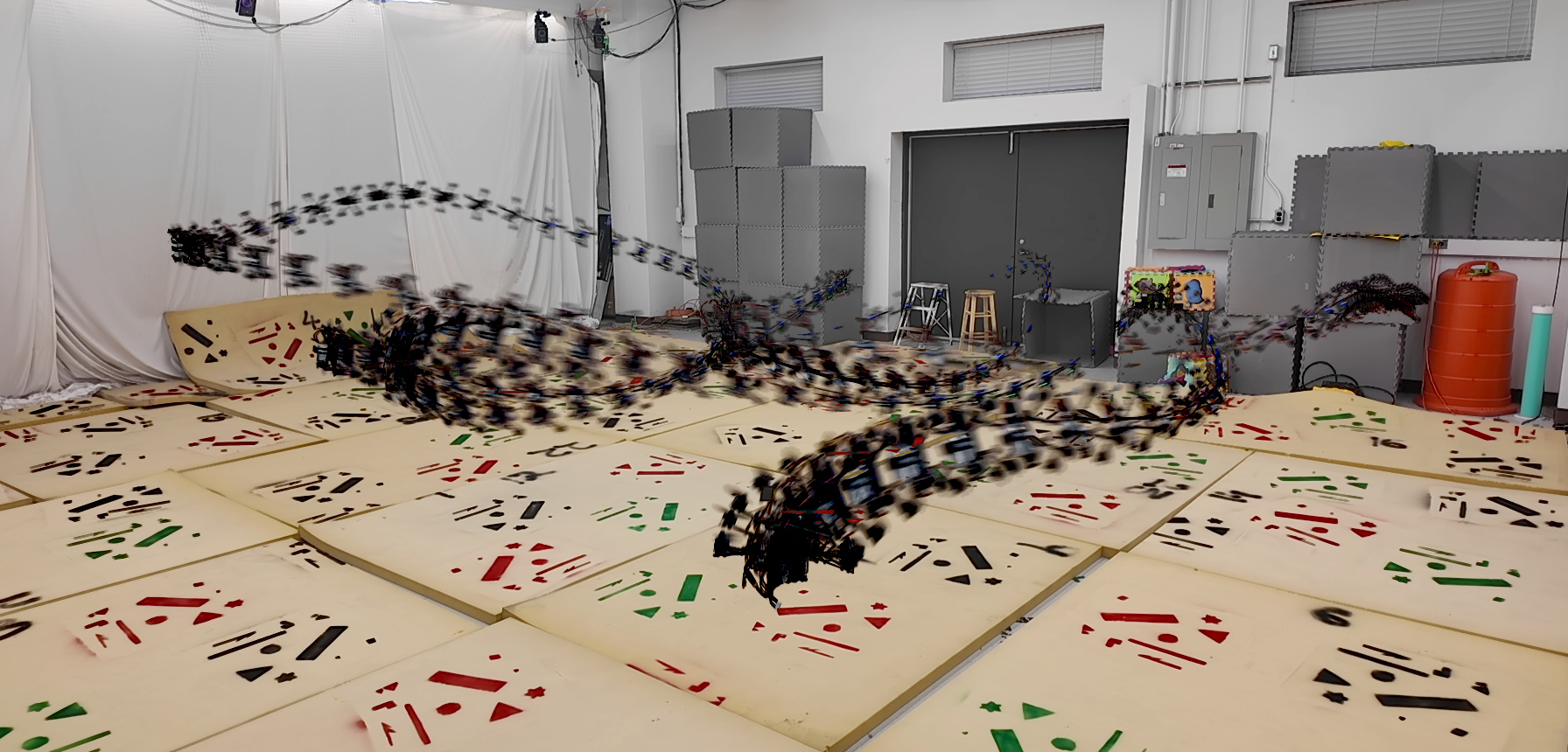}
\caption{Composite of an indoor high-speed flight.}
\label{fig:ghost}
\end{figure}

\begin{figure}
\centering
\includegraphics[width=\columnwidth]{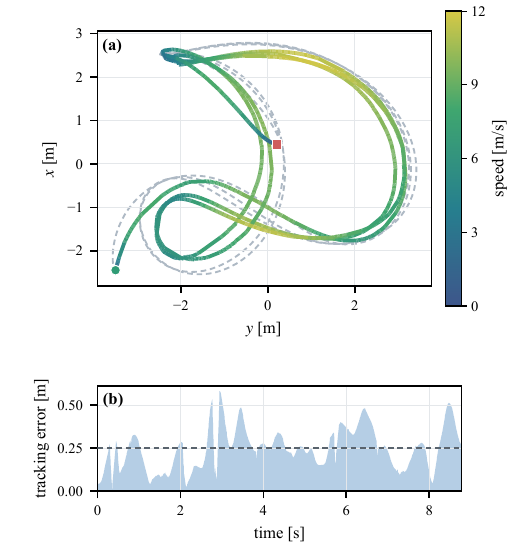}
\caption{Indoor time-optimal Split-S flight. (a) Planned path (gray dashed) and executed
trajectory, colored by estimated speed. (b) Three-dimensional tracking error.
The dashed line denotes the mean.
The planned path is 73.7\,m; the vehicle reaches
11.90\,m/s with 0.278 m root-mean-square tracking error.}
\label{fig:indoor}
\end{figure}

\begin{figure}
\centering
\includegraphics[width=\columnwidth]{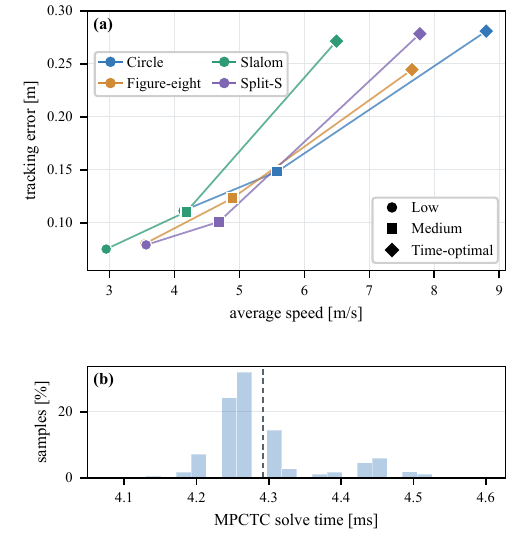}
\caption{Indoor operating envelope and onboard MPCTC solve time. (a) Average
executed speed and root-mean-square tracking error across four path geometries and three aggressiveness levels.
Connected markers denote the same geometry. (b) Distribution of MPCTC solve times; the dashed line denotes the mean.}
\label{fig:envelope}
\end{figure}

We first fly circle, figure-eight, slalom, and Split-S paths at three
aggressiveness levels. The vehicle reaches 12.38\,m/s with a root-mean-square
tracking error of 0.153\,m. Fig.~\ref{fig:envelope}(a) describes the correlation between speed and tracking error. Fig.~\ref{fig:ghost} shows one flight in the image plane,
while Fig.~\ref{fig:indoor} reports the time-optimal Split-S trajectory and
tracking error.

We schedule MPCTC every 10\,ms. Successful solves in
Fig.~\ref{fig:envelope}(b) have a 4.29-ms mean, 4.27-ms
median, and 4.55-ms
maximum.

\subsection{Outdoor validation}

We then move outdoors by replacing MoCap with RTK GNSS;
reference generation and control remain unchanged. The three-lap reference in
Fig.~\ref{fig:teaser} covers 438.6\,m in 21.49\,s over a
$32.2\times36.7$\,m footprint. With the complete stack
onboard, the vehicle reaches 31.4\,m/s (113.04\,km/h).

\section{Discussion and Conclusion}

Cybflight makes the flight controller firmware itself part of the research workflow rather than
a fixed layer beneath it. Its typed interfaces admit different hardware,
estimators, references, and controllers, while build-time selection specializes
the program for the target microcontroller. Our indoor and outdoor flights close
the complete perception-to-actuation path on one STM32H743. The results show that a modular Rust autopilot can execute
demanding estimation and nonlinear control fully onboard.

The design deliberately favors explicit, build-time composition. Changing a
module requires rebuilding and flashing the firmware, and fixed-size
\texttt{nalgebra} types catch matrix-dimension errors but not mistakes in
physical units or coordinate frames. We therefore keep those conventions
explicit at module boundaries and verify each configuration on the vehicle.
Within these constraints, a new method remains local to its module instead of
requiring changes throughout the flight stack. This modular foundation opens
three immediate paths for future research.

\textbf{Embedded learned control.} Direct-to-motor neural policies already
show the potential of running learned control on a flight
microcontroller~\cite{bahnam2026monorace}. We will connect neural-network
policies through the same typed state, reference, and actuator interfaces as
model-based controllers, so researchers can introduce a learned policy without
reorganizing perception and estimation.

\textbf{Swarm flight.} Crazyswarm showed that onboard computation and
low-bandwidth communication can scale to dozens of aerial
vehicles~\cite{preiss2017crazyswarm}. We will add typed peer-state and
coordination interfaces, allowing each vehicle to keep its perception,
estimation, and control onboard while sharing only the information needed for
coordination.

\textbf{Offboard control.} Some algorithms will still exceed the flight
microcontroller's computational budget. Aggressive predictive controllers,
for example, often run on larger processors~\cite{krinner2024mpccpp}. We will add
transport adapters that preserve the same state, reference, and command types,
so an algorithm can move between the flight microcontroller, a companion
computer, and offboard hardware without rewriting the surrounding experiment.


\begin{thebibliography}{99}
\bibitem{meier2015px4}
L. Meier, D. Honegger, and M. Pollefeys, ``PX4: A node-based multithreaded
open source robotics framework for deeply embedded platforms,'' in
\emph{Proc. IEEE ICRA}, 2015, pp. 6235--6240.

\bibitem{dangelo2024px4}
S. D'Angelo, F. Pagano, F. Longobardi, F. Ruggiero, and V. Lippiello,
``Efficient development of model-based controllers in PX4 firmware: A
template-based customization approach,'' in \emph{Proc. IEEE ICUAS}, 2024,
pp. 1155--1162.

\bibitem{bosello2024race}
M. Bosello \emph{et al.}, ``Race against the machine: A fully-annotated,
open-design dataset of autonomous and piloted high-speed flight,''
\emph{IEEE Robot. Autom. Lett.}, vol. 9, no. 4, pp. 3799--3806, 2024.

\bibitem{bahnam2026monorace}
S. A. Bahnam \emph{et al.}, ``MonoRace: Winning champion-level drone racing
with robust monocular AI,'' arXiv:2601.15222, 2026.

\bibitem{blaha2024indiflight}
T. M. Blaha, E. J. J. Smeur, and B. D. W. Remes, ``Control of unknown
quadrotors from a single throw,'' in \emph{Proc. IEEE/RSJ IROS}, 2024,
pp. 10350--10355.

\bibitem{foehn2022agilicious}
P. Foehn \emph{et al.}, ``Agilicious: Open-source and open-hardware agile
quadrotor for vision-based flight,'' \emph{Science Robotics}, vol. 7,
no. 67, Art. no. eabl6259, 2022.

\bibitem{romero2022mpcc}
A. Romero, S. Sun, P. Foehn, and D. Scaramuzza, ``Model predictive
contouring control for time-optimal quadrotor flight,'' \emph{IEEE Trans.
Robot.}, vol. 38, no. 6, pp. 3340--3356, 2022.

\bibitem{krinner2024mpccpp}
M. Krinner, A. Romero, L. Bauersfeld, M. Zeilinger, A. Carron, and
D. Scaramuzza, ``MPCC++: Model predictive contouring control for time-optimal
flight with safety constraints,'' in \emph{Proc. RSS}, 2024.

\bibitem{nguyen2024tinympc}
K. Nguyen, S. Schoedel, A. Alavilli, B. Plancher, and Z. Manchester,
``TinyMPC: Model-predictive control on resource-constrained
microcontrollers,'' in \emph{Proc. IEEE ICRA}, 2024.

\bibitem{kazim2023embedded}
M. Kazim \emph{et al.}, ``Aggressive trajectory tracking for nano
quadrotors using embedded nonlinear model predictive control,''
arXiv:2312.01015, 2023.

\bibitem{engels2025forces}
T. Engels \emph{et al.}, ``FORCES: An incremental transpiler from C/C++ to
Rust for robust and secure robotics systems,'' presented at the
\emph{ICRA Workshop on Rust for Robotics}, 2025.

\bibitem{schulz2025rio}
J. Schulz-Andres, H. Hose, and S. Trimpe, ``Robust radar--inertial odometry
with Rust: A seamless integration from embedded to high-level robotics,''
presented at the \emph{ICRA Workshop on Rust for Robotics}, 2025.

\bibitem{honig2026rusty}
W. H\"onig, C. Scherer, and K. Wahba, ``Rusty flying robots: Learning a full
robotics stack with real-time operation on an STM32 microcontroller,''
arXiv:2604.00032, 2026.

\bibitem{crozet2026nalgebra}
S. Crozet, ``nalgebra: General-purpose linear algebra for Rust,''
ver. 0.34.2, Mar. 2026. [Online]. Available:
\url{https://docs.rs/nalgebra/0.34.2/}

\bibitem{smeur2016indi}
E. J. J. Smeur, Q. P. Chu, and G. C. H. E. de Croon, ``Adaptive incremental
nonlinear dynamic inversion for attitude control of micro air vehicles,''
\emph{J. Guid. Control Dyn.}, vol. 39, no. 3, pp. 450--461, 2016.

\bibitem{qin2024togt}
C. Qin, M. S. J. Michet, J. Chen, and H. H.-T. Liu, ``Time-optimal
gate-traversing planner for autonomous drone racing,'' in
\emph{Proc. IEEE ICRA}, 2024, pp. 8693--8699.

\bibitem{qin2026ijrr}
C. Qin, J. Chen, Y. Lin, A. Goudar, A. Schoellig, and H. Liu,
``Time-optimal quadrotor maneuver: From optimality analysis to trajectory
planning,'' \emph{Int. J. Robot. Res.}, Jul. 2026.

\bibitem{preiss2017crazyswarm}
J. A. Preiss, W. H\"onig, G. S. Sukhatme, and N. Ayanian, ``Crazyswarm: A
large nano-quadcopter swarm,'' in \emph{Proc. IEEE ICRA}, 2017,
pp. 3299--3304.

\end{thebibliography}
\end{document}